\documentclass{svproc}
\usepackage{graphicx} 
\usepackage{algorithm}
\usepackage{algorithmicx}
\usepackage{amsmath}
\usepackage{amsfonts}
\usepackage{algpseudocode}
\usepackage{subfig}
\usepackage{bm}
\usepackage{multirow}
\usepackage[normalem]{ulem}
\useunder{\uline}{\ul}{}
\title{\LARGE \bf
Trade-offs of Dynamic Control Structure in Human-swarm Systems
}

\author{Thomas G. Kelly \and Mohammad D. Soorati \and Klaus-Peter Zauner \and
    Sarvapali D. Ramchurn \and and Danesh Tarapore}

\authorrunning{T.G. Kelly et al.}

\institute{School of Electronics and Computer Science, University of Southampton, Southampton, UK\\
\email{\{tgk2g14,mds1u19,kpz,sdr1,dst1m17\}@soton.ac.uk}}

\begin{document}

\maketitle
\thispagestyle{empty}
\pagestyle{empty}


\begin{abstract}

Swarm robotics is a study of simple robots that exhibit complex behaviour only by interacting locally with other robots and their environment. The control in swarm robotics is mainly distributed whereas centralised control is widely used in other fields of robotics. Centralised and decentralised control strategies both pose a unique set of benefits and drawbacks for the control of multi-robot systems. While decentralised systems are more scalable and resilient, they are less efficient compared to the centralised systems and they lead to excessive data transmissions to the human operators causing cognitive overload. We examine the trade-offs of each of these approaches in a human-swarm system to perform an environmental monitoring task and propose a flexible hybrid approach, which combines elements of hierarchical and decentralised systems. We find that a flexible hybrid system can outperform a centralised system (in our environmental monitoring task by $19.2\%$) while reducing the number of messages sent to a human operator (here by $23.1\%$). We conclude that establishing centralisation for a system is not always optimal for performance and that utilising aspects of centralised and decentralised systems can keep the swarm from hindering its performance.

\end{abstract}

\section{Introduction}

A swarm system presents a unique opportunity to exploit a huge number of simple agents and the redundancy that comes with it to complete tasks that would be impossible to solve by a single agent. Often, a decentralised approach is proposed as a scalable and resilient method for organising swarms but with low efficiency~\cite{heinrich2019}. On the contrary, centralised coordination can be efficient (e.g. in area coverage~\cite{karapetyan1846efficient}) but a centralised system is often not scalable and they rely on central units to control the system which can be points of system failure (i.e., low resilience). A combination of centralized and decentralized control could provide a swarm control method that makes use of the benefits of both approaches. Such a method is an important aspect of the future of robot swarms \cite{dorigo2020reflections}. Recently, a method inspired by Mergeable nervous systems \cite{mathews2017mergeable,zhu2020formation,jamshidpey2021centralization} was proposed that enables a swarm system to overcome the drawbacks faced by both centralised and decentralised systems. In this approach, a swarm is controlled through a self-organized hierarchical structure consisting of a brain robot which can be dynamically selected. A system using this approach is able to vary the level of decentralisation or centralisation according to mission requirements and for a given task. The relative advantages of using such an approach have been examined for varying groups of sizes of robots in both homogeneous \cite{mathews2017mergeable,zhu2020formation} and heterogeneous \cite{jamshidpey2020multi,jamshidpey2021centralization} systems.

Work considering theoretical foundations of self-organised hierarchical control frameworks has also been considered by Zhang et al. \cite{zhang2021self}. Hierarchical swarm control, despite being a relatively new field within swarm robotics, represents a promising area of research with important implications for the future of swarms \cite{dorigo2020reflections}. The self-organised nervous system for swarms was proposed by Zhu et al. \cite{zhu2024self} as a hierarchical control method for swarms, examining the scalability and stability of such a system. Work in this area, so far has focused on the application of a mergeable nervous system to distribute and coordinate a swarm based on a set of network topologies. While the relative performance of a bio-inspired hierarchical control structure has been studied, the performance of a more generic hybrid approach and its effect on human operators are not yet explored. When operating a swarm, the volume and quality of information exchanged with an operator and within the swarm is a key factor in maintaining situational awareness, and managing multiple tasks or interruptions at once increases the effort required by an operator to restore their situational awareness \cite{hussein2022characterization}. The effect of communication quality on swarm performance has been explored with communication aware approaches being proposed for hierarchical~\cite{xu2021communication} and decentralised~\cite{kelly2022collective} swarm control structures. To aid with the operation and situational awareness, the focus has been placed on developing novel interfaces for visualisation and control of swarms \cite{nagi2014human,patel2019mixed,divband2021designing}. To our knowledge, however, there is a lack of literature surrounding considerations of situational awareness with solutions that are tied to the swarm behaviour itself. Abioye et al.~\cite{abioye2023effect} studied the effect of cognitive workload on the overall human-swarm performance and showed that a higher volume or even quality of information sent to the operator does not necessarily increase the number of accurately completed tasks and slows down the operation.

Against this background, our paper aims at understanding how a simple hybrid control approach compares to either of the two (central or decentralised) in the overall system performance, the volume of messages sent to the operator (associated to cognitive workload) and the number of messages passed within the swarm. We study the trade-offs in using various levels of centralised coordination in an environment monitoring application where tasks appear randomly and the agents must move to the task area as soon as possible. We consider the human-swarm interaction implications intrinsically within the swarm behaviours. The hybrid approach is compared against a fully decentralised approach and a hierarchical approach in which agents act as a decentralised system or a hierarchical system depending on the state of the environment. Previous literature has examined the use of communication within a hybrid control approach but, to our knowledge, never from the perspective of a human swarm system and how such communications could potentially affect a human in the loop. We use a tree structure as an example of central control that observes events in the environment. We extend previous work done using photomorphogenesis for robot self-assembly \cite{divband2019photomorphogenesis} by allowing the formations to reconfigure themselves depending on the environment via the swarm behaviour. A recent study confirms that the swarm control strategy can be modified to reduce the cognitive workload~\cite{paas2022towards}. We, therefore, apply a dynamic hierarchical control approach and study the communication and the performance trade-offs that exist in using such a hybrid centralized and decentralized approach.  


We examine the assumption that a centralised approach always achieves the best performance in a task over a decentralised or hybrid solution. To do this, we propose a simple generic hybrid approach which combines elements of decentralised and centralised control systems. Due to its simple nature, it may be used for similar studies that wish to build on top of our work to further examine the costs and benefits of hybrid systems. We show that a hybrid system may achieve higher performance in some tasks (e.g., in our environmental monitoring task) compared to a centralised and decentralised approach. While this performance comes with trade-offs in communications, both between a human operator and within the swarm, we show that these trade-offs can be balanced with the performance of the system according to end-user requirements while using our proposed hybrid approach.

\section{Hybrid coordination}
\label{section:hier_coord}

A simple environmental monitoring task is used to assess each approach. Here, agents are tasked with observing and completing tasks in the environment. In order to complete an event, agents must observe the events for a set period of time, in our experiments this is 200 seconds. In our proposed hierarchical coordination algorithm, agents of the swarm dynamically form a forest of trees. The tree structure provides the end-user with a hierarchical control architecture for the swarm. Moreover, the trees are continually reshaped in response to changes in the operating environment. Both the task and the coordination we propose are simple in the interest of ensuring our approach is easy to follow and reproducible. We use the idea of tree formations as a basis for a centralised hierarchical approach.
\begin{figure}
    \centering
    \minipage{\textwidth}
    \subfloat[Decentralised]{\includegraphics[trim={2cm 1.3cm 1.6cm 1.4cm},clip,width=0.33\linewidth]{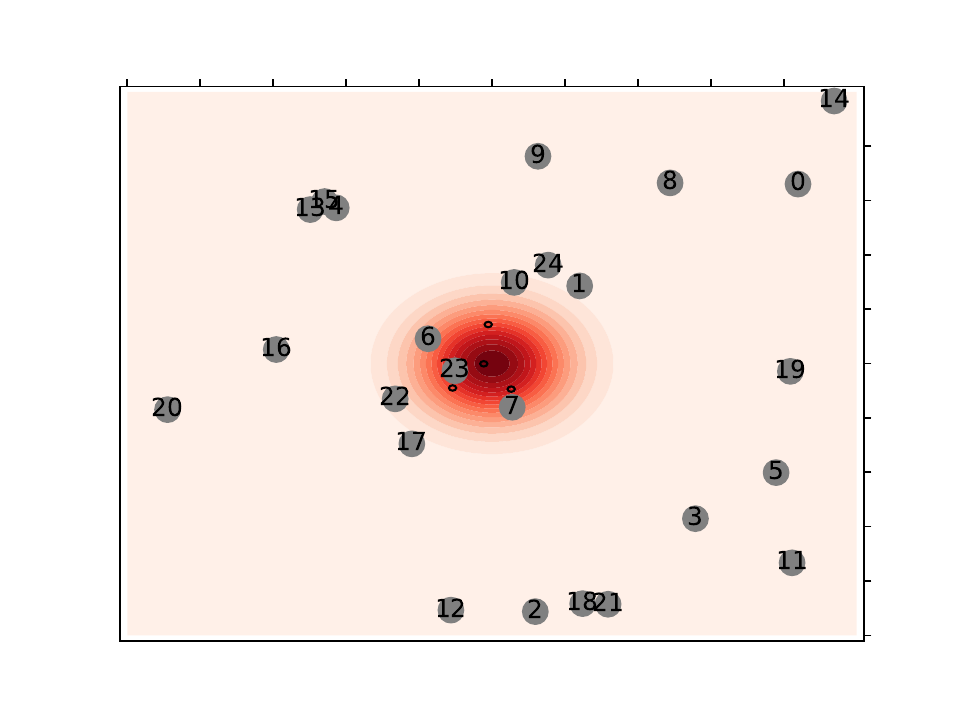}}
    \subfloat[Hierarchical]{\includegraphics[trim={2cm 1.3cm 1.6cm 1.4cm},clip,width=0.33\linewidth]{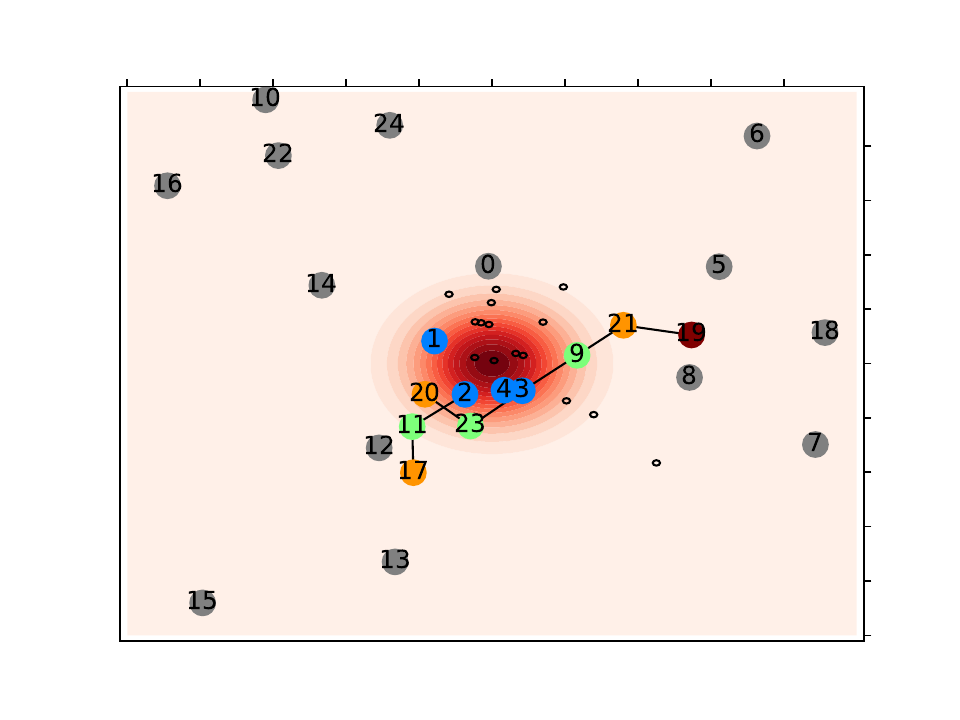}}
    \subfloat[Hybrid $\delta=3$]{\includegraphics[trim={2cm 1.3cm 1.6cm 1.4cm},clip,width=0.33\linewidth]{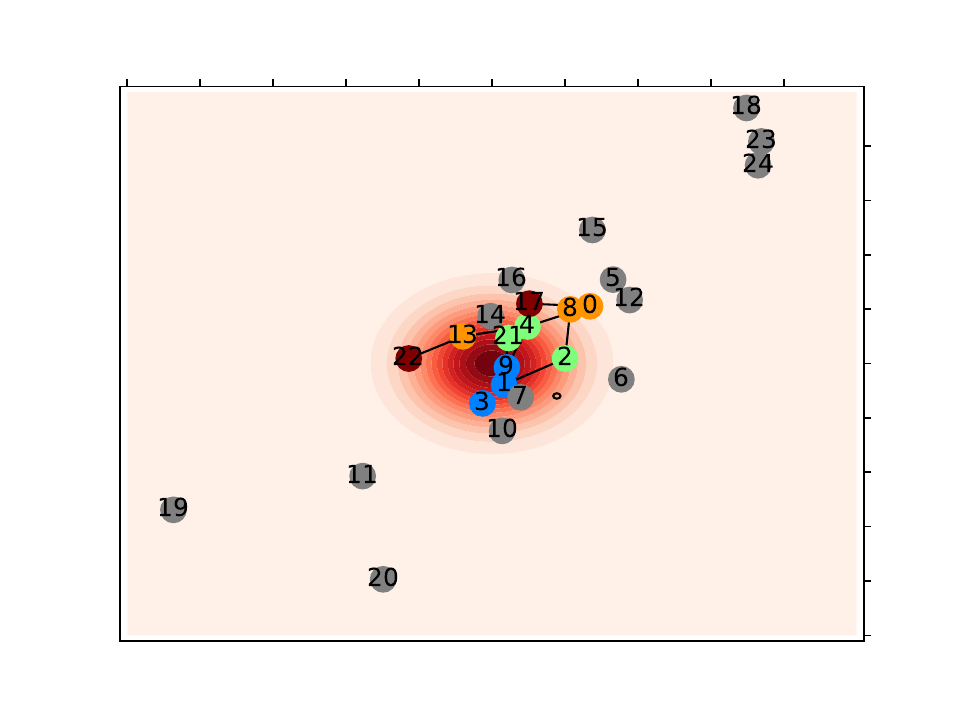}}
    \endminipage
    \caption{Examples the decentralised approach and early tree formations of the hierarchical and hybrid approaches during the task. The different coloured agents represent members of a tree at different depths, with root nodes being coloured blue. The small black circles represent existing incomplete events and the underlying event density distribution is shown.}
    \label{fig:examples}
\end{figure}
For our algorithm, we define a set of $n$ agents in the environment as $R$, with individual agents given as $r_i \in R$. Each agent, $r_i$, in a tree keeps track of its parent, its current level in the tree, $t_i$, and also a list of its followers, $F_i$, which is the set of descendants of agent $r_i$. An agent $r_i$, is said to be in a tree if it is a follower of some other node or a root, $\exists j \in \{1,...,|R|\} \wedge j \neq i$ s.t. $r_i \in F_j \lor r_i \in L$ where $L$ is the set of roots in the environment which is initially $L = \emptyset$. An agent becomes a root when that agent forms a new tree. When this happens, that agent is not part of any other tree and initially will have no followers. Agents share their lists of followers when they are within communication range of each other, and parents will aggregate these lists to create a list of all of their descendants, $F_j \rightarrow F_j \cup F_i$ where $r_j$ is the parent of $r_i$. 

When an agent, $r_i$, initially joins a tree, $F_i = \emptyset$. Each agent also maintains a set of scores, for each of its descendants, $S_i = \{s_j \forall r_j \in F_i\}$, which could relate to how well an agent is equipped to complete tasks that are occurring or present at its current location or to know how well the agent is positioned to maximise the swarm coverage.

During a mission, when an event occurs at a location $(x,y)$, $e_n^{(x,y)}$ where $n \in \{1,...,|E|\}$, the event will be added to the set of events currently in the environment that have not been completed by the swarm, $E$. Each agent stores a set of events that have been observed or communicated to them, $E_i \subseteq E$. This information is shared between agents when they are in communication range of each other. An agent also defines the set of all events that occur within the sensing range of a location, $(x,y)$ as $E_i^{(x,y)} \subseteq E_i$. 

When an event occurs in the environment and is observed by an agent in a tree, this information is passed up the tree to the root. This root aggregates all the information about events that are currently being observed by agents in the tree and communicates this as a message to an operator.

Our hierarchical coordination algorithm consists of four elements and aims to form a forest of trees of agents that can act as teams in areas of high activity and importance within the environment: 

\begin{itemize}
    \item Root Formation: This defines where a tree should be centred and aims to decide where a tree would be beneficial for the mission.
    \item Tree Growth: Adding resources to the tree in the form of other agents, to improve the coverage and detect events in that area.
    \item Active Recruitment: Reallocating the resources to the most needed areas and directing the growth of the tree to the most appropriate areas of the environment.
    \item Dissolution: The pruning of unnecessary branches of the tree, so the resources can be put back into the environment for better use.
\end{itemize}

Each of the four components of our coordination algorithm works as follows:

\vspace{.1cm}
\noindent\textbf{Root Formation:} An agent, $r_i$ forms a root node when the density of events currently occurring within the sensing range of the agent at its current location, $(x,y)$, exceeds an event density threshold, $\lvert E_i^{(x,y)} \rvert \geq \rho$. When this condition is satisfied, the agent stops moving, declares itself the root node for a new tree at its current location, such that $r_i \in L$. Initially, the set of followers for $r_i$ is empty, $F_i = \emptyset$.

\vspace{.1cm}
\noindent\textbf{Tree Growth:} Free agents, $r_i$, are members of the swarm that are not currently part of any tree, such that $r_i \notin F_j \mbox{ } \forall j \in \{1,...,|R|\} \wedge j \neq i$. They move around the environment by performing a custom random walk where agents move in a straight line for a uniformly distributed amount of time before turning onto a new randomly chosen heading. This method of movement was chosen to give agents towards the periphery of the environment a better chance of moving towards the area of interest where events were occurring. When a free agent moves within communication range of another member of the swarm, $r_j$, that is in a tree, such that $\lvert \overrightarrow{r_i r_j} \rvert \geq d_{min}$ so the two agents are at least $d_{min}$ meters apart, then $r_i$ joins the tree, with $r_j$ as its parent and $r_i \in F_j$.

\vspace{.1cm}
\noindent\textbf{Active Recruitment:} Periodically, in our experiment once every minute, root nodes perform active recruitment in order to improve the performance of the tree. This is done to relocate any agents that are currently part of the tree at locations where events are not as frequent as other areas of the environment. It also gives the tree a chance to expand into areas that the root feels are more important to detect a greater number of events than other areas of the environment. The root, $r_j$ first determines the agent that will be moved in order to improve the number of events or expected number of events monitored by the tree. This is done by selecting a leaf node, $r_i$, of the tree which is contributing the least, in terms of events monitored, to the tree, such that $\displaystyle \min_{s_i \in S_j}\{s_i \lvert F_i = \emptyset\}$.

The root considers all possible locations, $(x,y)$, that agent $r_i$ can move to while remaining within the communication range of at least one of the other members of the tree, $r_k$, and also satisfying the condition for the tree growth, $\lvert \overrightarrow{r_i r_k} \rvert \geq d_{min}$. The root assigns a score to each location, $a_{(x,y)} = \lvert E_j^{(x,y)} \rvert$. The agent $r_i$ then moves to $(x,y)$, changing its parent as appropriate, once the new location is reached.

Since only leaf nodes are considered for active recruitment, there is never an issue  of breaking apart an existing branch of the tree, disrupting the overall tree. The aim of the active recruitment is to promote the growth of the tree into areas that have many events occurring.

\vspace{.1cm}
\noindent\textbf{Dissolution:} Agents that are in a tree maintain a dissolution parameter, $\gamma$, that is set to the maximum value, $\gamma_{max}$ when an event, $e_i^{(x,y)}$, is observed by the agent at their current location, which acts as a stimulus. This parameter is decreases linearlly at each timestep. The rate at which $\gamma$ decreases for a given agent is proportional to the number of parents above that agent in the tree, $t_i$, given by $\gamma = \gamma - t_i$. When $\gamma = 0$ for an agent, $r_i$, the agent disconnects itself from the tree and becomes a free agent again. As such, agents that are closer to the periphery of the tree will dissolve quicker than those closer to the root. This allows the tree to remain more dynamic, reducing the potential for having agents remain in unnecessary locations where few events are occurring, while allowing a tree with a well placed agents within the environment to persist. When an agent disconnects itself from the tree, that information is propagated to the rest of the tree by $r_i$'s root so each agent in the tree can update their beliefs of the tree, $F_j \setminus r_i \forall j \in \{1,...,|R|\}$.

This parameter is used not only to limit the size of trees but also to ensure that trees are forming in locations where events are occurring with a higher likelihood. If a tree does not detect any events for a long period, it is not necessary to maintain a large tree in that area. The algorithm would dissolve the tree and allow new trees to form in different areas.

\vspace{.1cm}
\noindent\textbf{Hybrid coordination:} The hybrid approach is a combination of the hierarchical approach and the decentralised approach. This is achieved by having agents act as a decentralised system or a hierarchical system, in response to the conditions in the environment. The level of centralisation or decentralisation is controlled by the parameter $\delta$ which corresponds to the number of events that an agent must be monitoring in order to switch from a decentralised to a centralised system.

When an agent, $r_i$ observes an event, at $(x,y)$, the robot will perfomr one of two actions. If the density of events currently occurring within sensing range of the agent at its current location exceeds the event density threshold, $\lvert E_i^{(x,y)} \rvert \geq \delta$, then the agent will act as a hierarchical system and form a root node. If this condition is not met, $\lvert E_i^{(x,y)} \rvert < \delta$, then the agent will behave as a decentralised system and will not form a root node but will change its heading to move towards the event and it will proceed to observe the event.

\vspace{.1cm}\noindent\textbf{Baseline coordination algorithm:} We compare our hierarchical and hybrid coordination algorithms against a standard decentralised approach. In the decentralised approach, agents explore the environment via a random walk and exchange information regarding the location and status of events that have been observed, as in the hierarchical approach. When an agent observes an event, the agent corrects its heading to move directly towards the event. Once the agent reaches the location of the event, it stops moving and observes the event until it is completed, when the agent continues moving randomly in the environment. The event is considered completed when it has been observed for 200 seconds, at which point that event is removed from the environment. If another event occurs within the agent's sensing range while it is attending to an event, then the agent will immediately move to the new event, once the original event that the agent was observing is complete.

If a randomly moving agent, which is not observing an event moves within communication range of an agent that is observing an event, then the randomly moving agent will, again, alter its heading to move towards the event and assist in the observation of the event. Agents also communicate information about events that they are observing to an operator.
\section{Experimental Setup}
Agents move at a maximum speed of 1 m/s in a bounded environment, $Q \subseteq \mathbb{R}$ which is discretized into a set of points, $(x,y) \in Q$. For our experiments, we fix the density of agents in the environment to 0.0025 agents per $m^2$. So for the experiments with 25 agents, the environment is 100m $\times$ 100m with 20 replicates of 9000 seconds. We assume that there exists an unknown density function $\phi(x,y):Q \rightarrow \mathbb{R}^+$ which defines the probability that an event, $e_i^{(x,y)}$ occurs at location $(x,y)$. For our experiment, in the base scenario, the distribution takes the form of a Gaussian distribution, with mean at $(50,50)$ and covariance of $\left( \begin{smallmatrix} 50&0\\ 0&50 \end{smallmatrix} \right)$. Events are persistent in the environment, until they have been observed by agents a set number of times, when they are completed and removed from the environment. Each distinct agent that observes an event contributes towards the event's completion, reducing the number of times an event needs to be observed before it is completed by 1. As such, having multiple agents observe an event will result in the event being completed more quickly and this increase in speed is linear in relation to the number of agents observing the event. For our experiments, we do not consider collision avoidance.

We evaluate the proposed swarm coordination approaches using the following metrics: i. Waiting time for events before being observed by a member of the swarm; this metric shows how responsive the swarm is to changes in the environment and how well the swarm is able to cover areas where events are occurring with a high probability; and ii. Number of messages received by a human operator at each timestep from distinct agents; this gives a measurement for the cognitive load that an operator may experience while controlling the swarm. Messages are received when an agent is observing an event if it is operating as a decentralised agent and from the root of a tree where multiple agents may be monitoring events within the tree.

\section{Results}
\subsection{Performance of swarm coordination}
\begin{figure}
    \centering
    \minipage{\textwidth}
    \includegraphics[trim={0cm 0cm 0cm 0cm},clip,width=\linewidth]{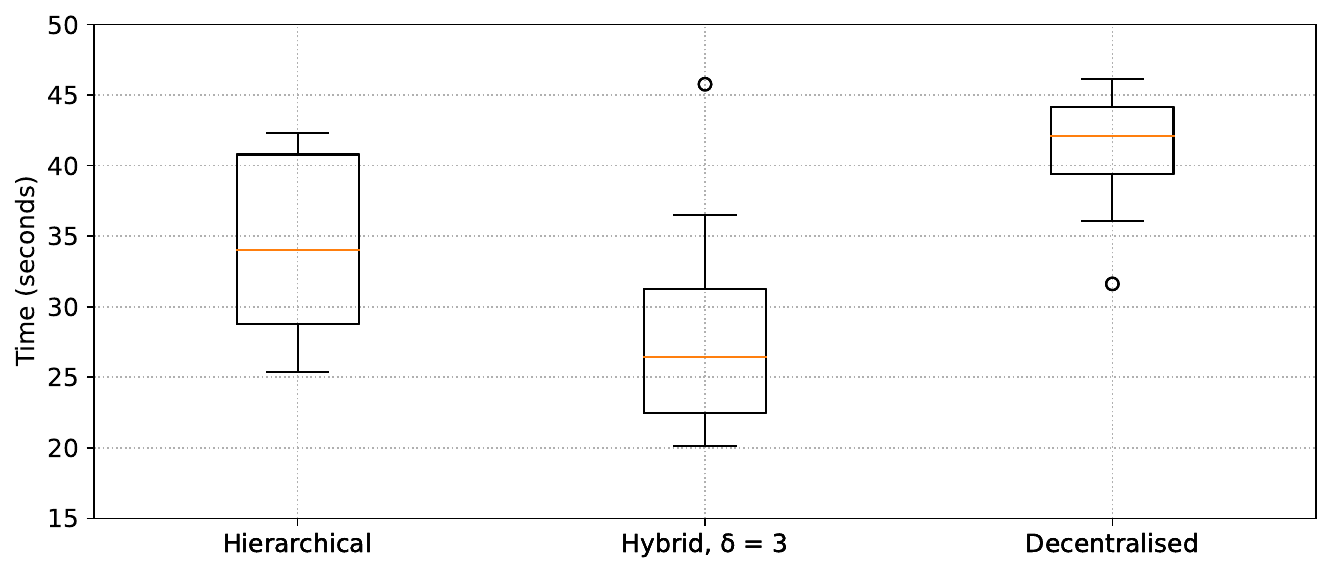}    
    \endminipage
    \caption{Box plot of waiting times for environmental events to be observed by a swarm of 25 agents coordinated with the Hierarchical, Decentralised and Hybrid approaches. Event waiting time data was averaged across each replicate, and aggregated across 20 replicates.}
    \label{fig:attendance_25}
\end{figure}

Figure~\ref{fig:attendance_25} shows the mean waiting times taken for events to be observed by members of a swarm of 25 agents coordinated with the Hierarchical, Hybrid and Decentralised approaches over 10 separate and independent replicates. The Hierarchical approach improves on the performance in monitoring environmental events over the decentralised approach with median$\pm$IQR of 42.1$\pm$4.7 seconds and 34.0$\pm$12.0 seconds for the Decentralised and Hierarchical approaches, respectively. This shows a $19.2\%$ reduction in the waiting time. The hybrid approach also outperforms both the Hierarchical and Decentralised approaches with median$\pm$IQR waiting time of 26.4$\pm$8.8 seconds, a $37\%$ reduction in waiting time over the decentralised approach and a $22\%$ reduction over the hierarchical approach.
The hybrid approach is able to improve the performance of the swarm for this task, over that of the hierarchical and decentralised approaches due to its ability to act as both a centralised and decentralised system. Furthermore, the value of $\delta$ dictates how much time an individual should spend as each of the approaches. As such, at the beginning of the experiment, agents with a higher $\delta$ value will remain flexible, being more likely to act as a decentralised system and not commit to forming trees until there is more certainty over the distribution of events. Alternatively, agents with lower $\delta$ values will form trees more quickly but may form them in sub-optimal locations which will take longer to adjust with the active recruitment. This can be seen in figure ~\ref{fig:examples}(b) and (c) where the hierarchical approach has initially formed trees away from the centre of the event density function, while the hybrid approach has trees that have formed closer to the centre. However, agents that have $\delta$ values that are too high and act as a decentralised system for a large proportion of time will not gain the benefits of coordination.

\subsection{Scalability of coordination algorithms}
\begin{figure}
    \centering
    \minipage{\textwidth}
    \includegraphics[trim={0cm 0cm 0cm 0cm},clip,width=\linewidth]{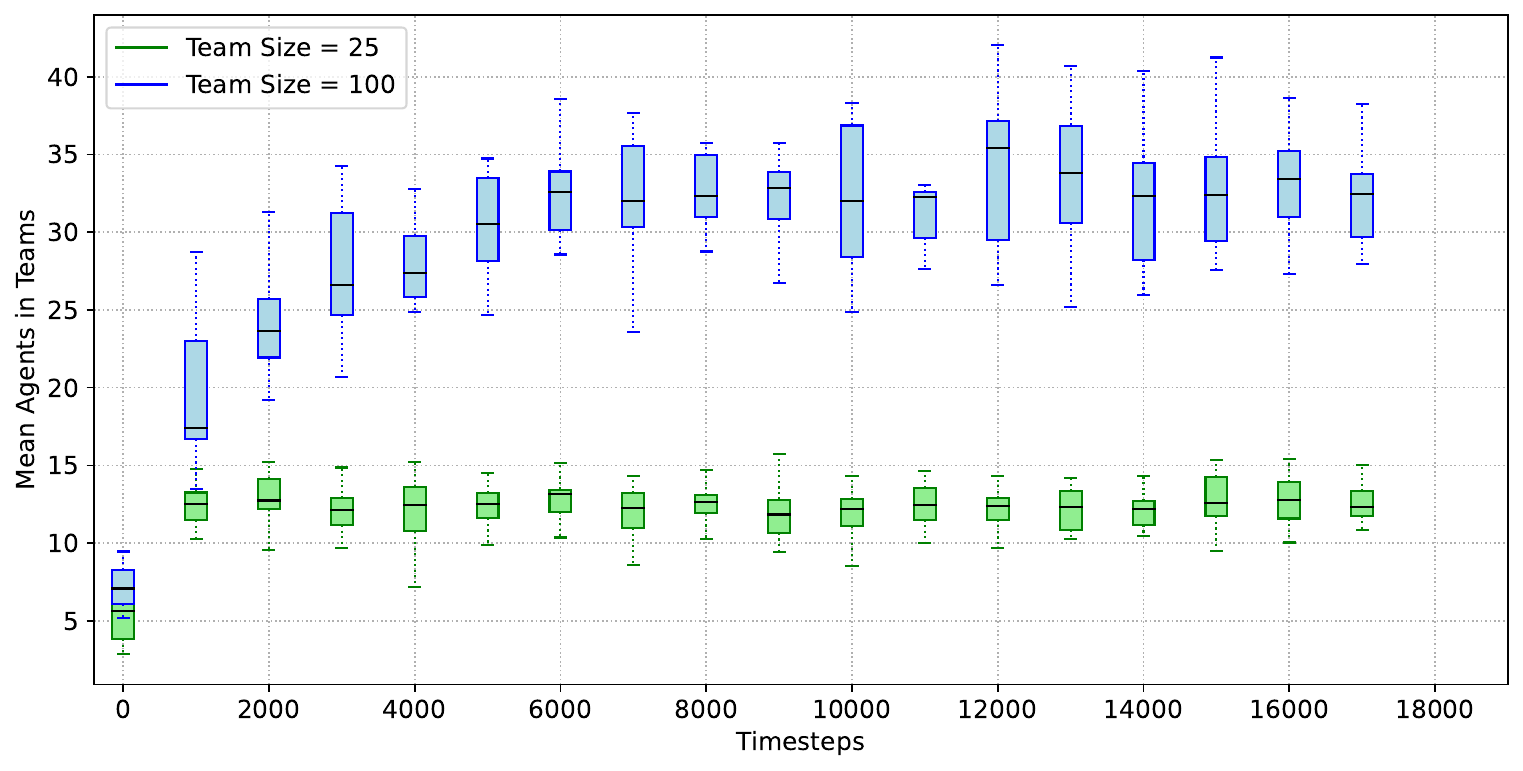}
    \endminipage
    \caption{Mean number of agents that are part of trees by swarms of sizes 25 and 100 agents coordinated with the hybrid approach with $\delta = 3$. Data was aggregated across 20 replicates.}
    \label{fig:scale_convergence}
\end{figure}

We examine the scalability of the hybrid approach by considering the number of agents present in trees over the course of the experiments. This is shown in Figure~\ref{fig:scale_convergence} for swarm sizes of 25 and 100 agents. In these experiments, the environment was scaled proportionally to the increase in the size of the swarm sizes in order to keep the density of agents present in the environment equal. The event density function and frequency of events were also scaled to account for the increase in the size of the environment and the swarm size. The swarm of size 25 quickly converges to having 12 agents present in teams at any time within the first 1000 timesteps, while the swarm size of 100 converges to 33 agents in teams at any point in time after 6000 timesteps. This difference in convergence time is likely due to the increase in the size of the environment and the event density function, while the sensing of the range of agents remains the same. As a result, agents spend more time behaving as a decentralised system initially. This is due to events having a greater chance of occurring over a larger area. This means that initially, the density of events in the environment do not exceed the swarm's $\delta$ threshold value. As such, trees do not begin to form for some time.

This shows the hybrid coordination algorithm remains stable and predictable for swarms with different sizes and for larger environments. This is validated further by considering the coverage of the two swarm sizes. The upper limit of the possible coverage by teams of size 12 and 33 agents, assuming a sensing range of 5m and no overlap, is $942$m$^2$ and $2591$m$^2$ respectively. The actual coverage of the swarms is less than these totals, with a mean coverage of $672$m$^2$ and $1521$m$^2$ for a swarm with a team of 12 agents and 33 agents respectively. These totals exceed the area of the environment where $95\%$ of the events occur. This portion of the event density function for a swarm of size 25 and 100 is a circular area of $628$m$^2$ and $1256$m$^2$ centred at the midpoint of the environment. As such, both swarm sizes are able to converge to teams that have the ability to cover $95\%$ of the events that occur in the environment.
\subsection{Ablation study}
\begin{figure}
    \centering
    \minipage{\textwidth}
    {\includegraphics[trim={0cm 0cm 0cm 0cm},clip,width=\linewidth]{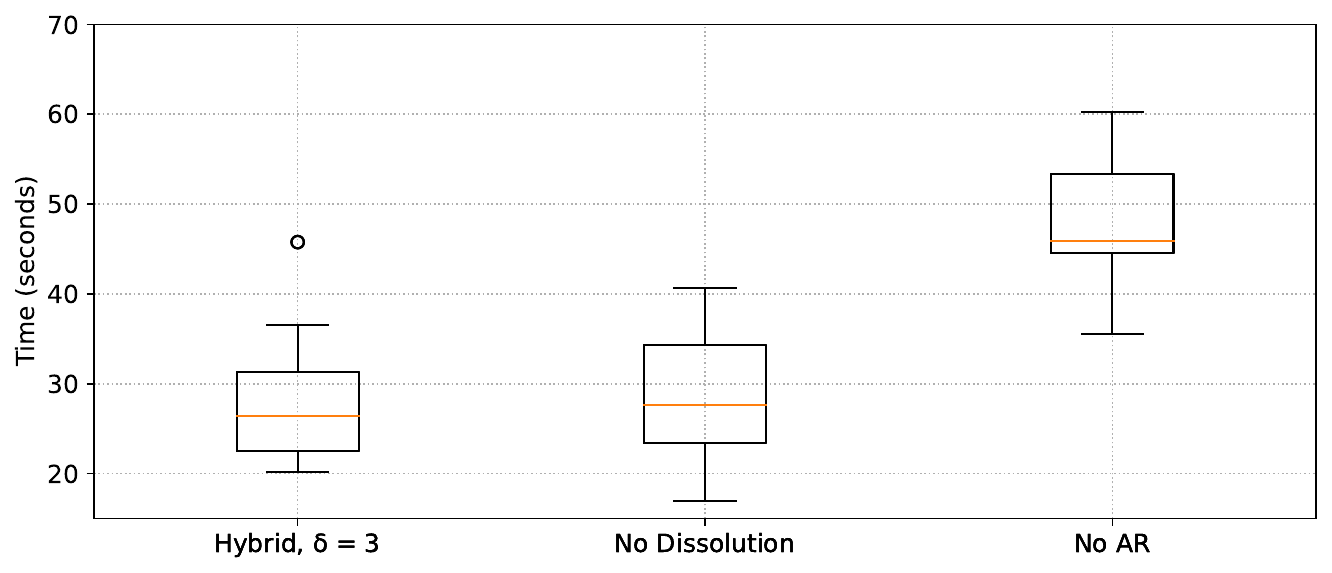}}
    \endminipage
    \caption{Ablation analysis of mean waiting times for events to be observed by a swarm of 25 hierarchically coordinated agents following the removal of the active recruitment (AR), and the dissolution elements.}
    \label{fig:ablation}
\end{figure}

Figure~\ref{fig:ablation} compares the performance of the hierarchical approach against the approach when there is no active recruitment and when there is no dissolution of the trees. The median$\pm$IQR waiting time for the hybrid approach is 26.4$\pm$8.8 seconds and we see a $73\%$ increase in the waiting time when the active recruitment is removed to 45.9$\pm$8.8 seconds. Even though the active recruitment only occurs periodically, it has an important effect on the performance. We also observe a $5\%$ increase in the waiting time when the dissolution of the trees is removed with a median$\pm$IQR time of 27.6$\pm$10.8 seconds. This suggests that both the dissolution and the active recruitment helps to keep the trees flexible, even when the other is not present, with the active recruitment affecting the performance more than the dissolution element which occurs more frequently.

It is clear that the coordination provided by the active recruitment for the Hierarchical and Hybrid approaches is an important factor in improving the performance over that of the decentralised approach. When there is no active recruitment present, the performance of the hierarchical approach, with a median$\pm$IQR waiting time of 45.9$\pm$8.8 seconds, is close to that of the decentralised approach, with 42.7$\pm$5.2 seconds.

\subsection{$\delta$ Sensitivity analysis}
\begin{figure}[h]
    \centering
    \minipage{\textwidth}
    {\includegraphics[trim={0cm 0cm 0cm 0cm},clip,width=\linewidth]{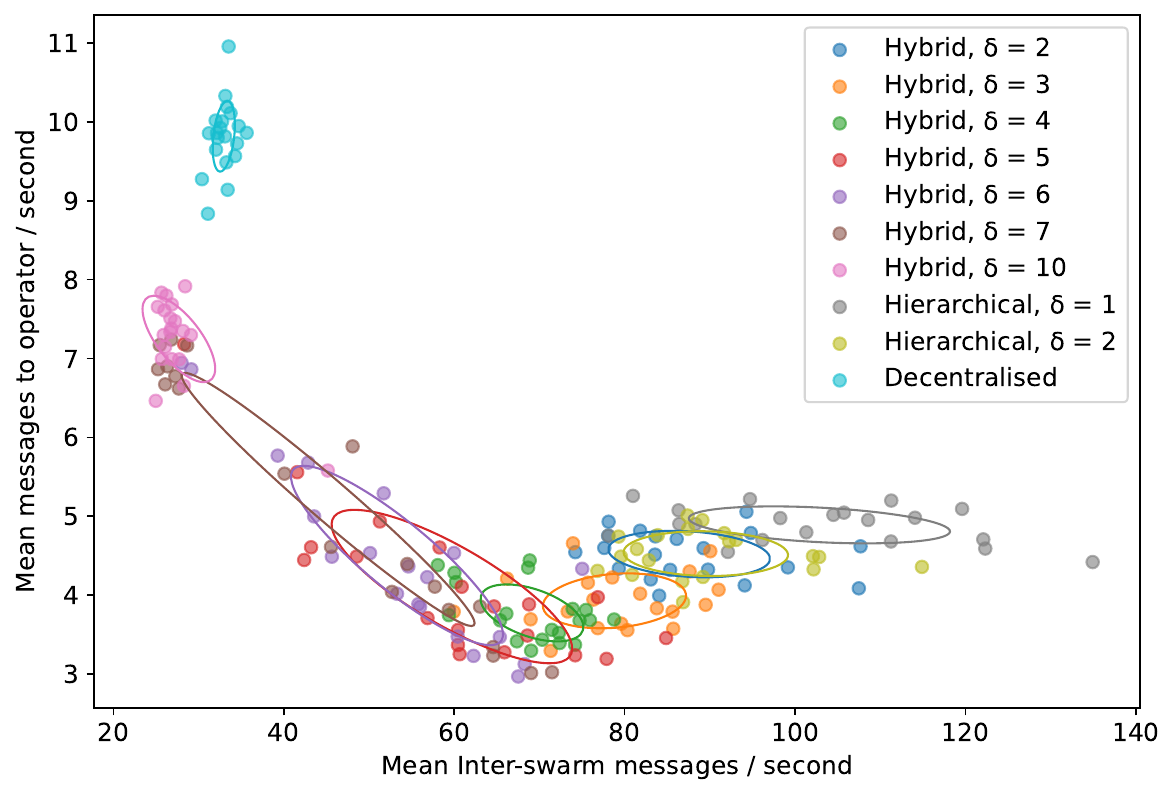}}     
    \endminipage
    \caption{Mean number of messages sent by different agents in a swarm of size 25 to a human operator against the mean number of messages sent at each timestep within the swarm.}
    \label{fig:comms_dependency}
\end{figure}

Figure~\ref{fig:comms_dependency} shows the mean number of messages sent to an operator per second against the number of inter-swarm messages (between members) per second over the course of 20 runs for a swarm of 25 agents. Here, we vary the levels of centralisation, these being decentralised, hierarchical and hybrid with $\delta$ values ranging from 3 (more centralised) to 10 (more decentralised). The points for each of the replicates are shown with ellipses that represent the mean and one standard deviation from the mean for each of the approaches. We can see that a Hybrid approach with a low $\delta$ value acts close to that of a hierarchical system and as the delta increases that hybrid system behaves closer to that of a decentralised system.

When we consider the cognitive load, we see that the hybrid $\delta = 3$ and hierarchical $\delta=2$ approaches, with median$\pm$IQR of 3.9$\pm$0.50 and 4.5$\pm$0.42 messages to an operator per second, reduce the number of messages sent to the operator by over 50\% compared to the decentralised approach, with median$\pm$IQR of 9.9$\pm$0.38 messages to an operator per second. The lower load on the operator overseeing the Hierarchical and Hybrid swarms is consequent to the information sent from the relatively few dynamically formed leaders of the trees has been aggregated; the multiple environmental events spanning the area covered by a tree do not require the communication of multiple messages as in the decentralised approach.

We see the best performance of the system for this scenario for $\delta = 3$ and $\delta = 4$. When a member of a Hybrid system senses events in the environment, it must check whether the density of events exceeds $\delta$ before deciding to behave as a decentralised system or hierarchical system by forming a root. Smaller values of delta increase the chance of an agent forming a root node since a lower density of events is required. This results in multiple trees being formed in sub-optimal locations within the environment where the density of events is lower than the maximum density present in the environment. This leads to more leaders being present in the environment and as a result more messages being sent to the operator. Larger $\delta$ values reduce the probability of a root forming since the required density of events is much greater. This means fewer trees form and more agents continue to operate as a decentralised system. If an appropriate delta value is chosen to reflect the maximum density of events present in the environment, then we see a minimal number of trees with roots focused near the points in the environment with the maximum density, allowing for fewer leaders and therefore fewer messages sent to the operator, along with a well placed tree to cover the area of interest.

This shows that centralised control does not always provide the best performance for a task. It suggests that a system which is too sensitive to stimuli that cause it to centralise can get trapped in local minima. In order to overcome this issue, a more flexible solution is required that refrains from centralisation too quickly. This would allow time to assess the scenario and adapt accordingly. While this sacrifices performance in the short term, it guarantees a better performance over time.

\section{Conclusions and future work}

We have analysed a control structure that gives flexibility over the level of centralisation or decentralisation for a robot swarm system. The performance of such a system and the effects the system could have on communication with a human operator in an environmental monitoring task have been evaluated. The results suggest that having a swarm system centralise too quickly can be detrimental to the performance of the swarm and result in the system becoming stuck in local minima. In order to overcome this, a swarm needs to remain flexible during its mission, ensuring not to centralise until enough information has been gathered about the environment. Finding a balance between centralised control and decentralised control can improve this, but comes at a cost of performance in the short term. Future work will consider how an aspect of learning could be added to the hybrid approach that will allow the swarm to adapt the delta value in response to the task and the environment. This would give the swarm control over its level of centralisation as needed.

The hierarchical and the hybrid approaches have both been shown to outperform the decentralised approach in an environmental monitoring task, while also reducing the number of messages sent to a human operator, thus providing a reduction in the noise received by that operator. One drawback to these approaches is that they come with inter-swarm communication and message size overheads. However, we have shown that by varying $\delta$ for the hybrid approach, different levels of centralisation can be achieved in order to find an optimal level of inter-swarm and swarm-to-operator communication for a given mission. These communication overheads pose another interesting question for future work in how these approaches might be affected by communication limitations in the environment. Whether this would affect the overall performance of the hierarchical and hybrid approaches could also be explored. Another direction for future work is to allow the swarm to select its $\delta$ value during runtime to adapt to the changing environment and based on different tasks. 

%
%
%
\bibliographystyle{splncs04}
\bibliography{root}
\end{document}